\documentclass[a4paper,conference]{IEEEtran}
\ifCLASSINFOpdf
   \usepackage[pdftex]{graphicx}
\else
   \usepackage[dvips]{graphicx}
\fi
\usepackage{booktabs}

\usepackage{amsmath}
\usepackage{url}
\begin{document}

\title{SIMCO: SIMilarity-based object COunting}

\author{\IEEEauthorblockN{Marco Godi\textsuperscript{\textsection}, Christian Joppi\textsuperscript{\textsection}, Andrea Giachetti, Marco Cristani}
\IEEEauthorblockA{Department of Computer Science\\
University of Verona\\
Verona\\
Email: [name].[surname]@univr.it}

}


%Telephone: (800) 555--1212, Fax: (888) 555--1212}


\maketitle
\begingroup\renewcommand\thefootnote{\textsection}
\footnotetext{Equal contribution}
\endgroup
\begin{abstract}
We present SIMCO, a completely agnostic multi-class object counting approach.
SIMCO starts by detecting foreground objects through a novel Mask RCNN-based architecture trained beforehand (just once) on a brand-new synthetic 2D shape dataset, InShape; the idea is to highlight every object resembling a primitive 2D shape (circle, square, rectangle, etc.). Each object detected is described by a low-dimensional embedding, obtained from a novel similarity-based head branch; this latter implements a triplet loss, encouraging similar objects (same 2D shape + color and scale) to map close. Subsequently, SIMCO uses this embedding for clustering, so that different ``classes'' of similar objects can emerge and be counted, making SIMCO the very first multi-class unsupervised counter. The only required assumption is that repeated objects are present in the image. Experiments show that SIMCO provides state-of-the-art scores on counting benchmarks and that it can also help in many challenging image understanding tasks. 
\end{abstract}


% \fi
\IEEEpeerreviewmaketitle

\section{Introduction}

Most approaches for counting similar objects in images assume a single object class~\cite{kang2018beyond, norouzzadeh2018automatically, onoro2016towards, falk2018u}; when is not, ad-hoc learning is necessary~\cite{lu2018class, setti2018count, laradji2018blobs}. 
None of them are truly agnostic and multi-class, i.e., able to capture generic repeated patterns of different type \emph{without any tuning}.
The work of Laradji~\cite{laradji2018blobs} can be applied to different datasets after completely training the model; in the work of Lu~\cite{lu2018class} it is not required to fully retrain the model, but a specialization phase that involves the training of an adapter module with a few labeled samples is still required.
The work of Setti~\cite{setti2018count} is the most similar one to this work, in the sense that no training is required at all.
Counting approaches are based on regression (\cite{Zhang_2015_CVPR}) or density estimation  (\cite{onoro2016towards, kang2018beyond}); 
here we focus on counting by detection~\cite{setti2018count, subburaman2012counting, laradji2018blobs} so the counted objects are individually detected first. %, in order to exploit their visual appearance afterwards. 

\begin{figure}[t!]
\includegraphics[width=8.0cm]{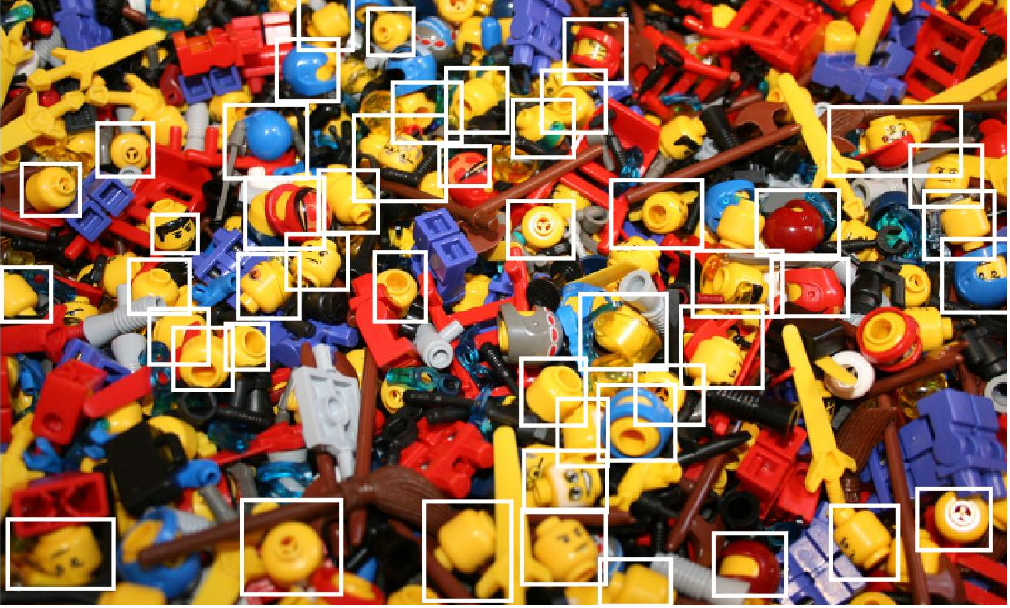}
  \caption{
\textbf{SIMCO on visual question answering}: the most occurrent object? SIMCO finds 47 LEGO heads. 
  } 
\label{fig:lego}

\end{figure}

Research on agnostic counting is important in many fields. It serves for visual question answering~\cite{jabri2016revisiting,zhang2018learning}, where counting questions could be made on too-specific entities, outside the semantic span of the available classes~\cite{agrawal2016analyzing} (e.g. ``What is the most occurrent thing?'' in Fig.~\ref{fig:lego}). In representation learning, unsupervised counting of visual primitives (i.e., visual ``things'') is crucial to obtain a rich image representation~\cite{onoro2016towards,noroozi2017representation,katsuki2016unsupervised}. Counting is a popular topic in cognitive  robotics~\cite{rucinski2014modelling,cangelosi2016embodied}, where autonomous agents learn by separating sensory input into a finite number of classes (without a precise semantics), building a classification system that counts on each %one 
of them. 

Application-wise, agnostic counting may help the manual tagging of training images ~\cite{andriluka2018fluid}, providing a starting guess for the annotator on  single-~\cite{spampinato2008detecting,joly2016lifeclef} or multi-spectral~\cite{ferrari2017bacterial,hollings2018you} images. %Other potential applications of SIMCO are to find the most occurrent object in a scene (Fig. ~\ref{fig:lego})... %
Inpainting filters on image editing applications may benefit from a magic wand capturing repeated instances to remove. Examples of these applications are shown in Sec. \ref{sec:experandresult}.
These kinds of tasks require a framework that works \emph{without any kind of training/specialization phase}, since training data may not be available (e.g. a single image of that kind in which to perform counting or a small set of images without any annotations).

In this paper, we present the SIMilarity-based object COunting (SIMCO) approach, which is completely agnostic, i.e. with no need of any \emph{ad-hoc} class-specific fine tuning, and multi-class, i.e. finding different types of repeated patterns. The method works on the assumption that the image contains repeated elements and those elements are the only ones that can be counted (outlier elements that are not repeated, are filtered away). It is a reasonable assumption, since it is usually desirable to count multiple repeated elements.
%se serve spazio mettere like
Two main ideas characterise SIMCO.  

First, every object to be counted is considered as a specialization of a %precise
basic 2D shape: this is particularly true with many and small objects
%, as in the case of counting problems %%cosa voleva dire questo?
~\cite{rosch1976basic,yan2017intelligent} (see Fig.~\ref{fig:lego}: LEGO heads can be approximated as circles). 
%, selected from a proper 2D shape ontology~\cite{rovetto2011shape,niknam2011modeling};
SIMCO incorporates this idea building upon a novel Mask-RCNN-based classifier, fine-tuned just once
on a novel synthetic shape dataset, \emph{InShape}. 

The second idea is that one can perform unsupervised grouping of the detected objects, discovering %not just one but 
\emph{different} types of repeated entities (without resorting to a particular set of classes). SIMCO realizes this with a head branch in the network architecture implementing triplet losses, which provides a 64-dim embedding that maps objects close if they share the same shape class plus some appearance attributes. Affinity propagation clustering~\cite{frey2007clustering} finds groups over this embedding.  
Results are state-of-the-art on cell counting benchmarks~\cite{arteta2014interactive}, and on the RepTile dataset~\cite{setti2018count}, explicitly suited for agnostic object counting. Qualitative results illustrate some of the many scenarios where SIMCO can definitely help.

\section{The SIMCO approach}
\label{sec:SIMCO}
\begin{figure*}[t!]
  \begin{center}
  \includegraphics[width=0.98\textwidth]{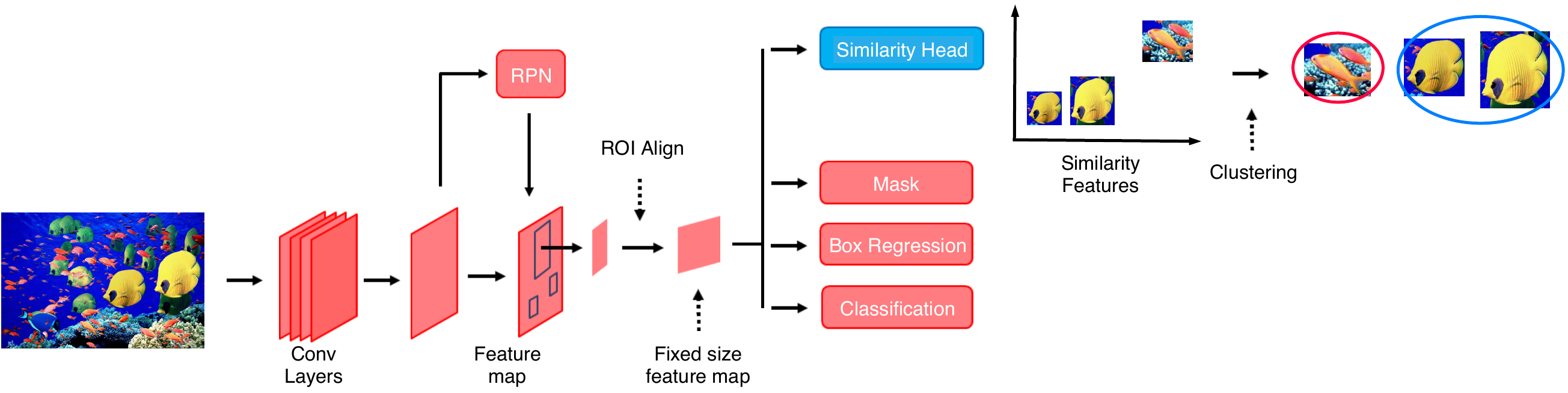}
  \end{center}
  \vspace{-1em}
  \caption{
\textbf{SIMCO scheme.} In red the standard Mask R-CNN~\cite{he2017mask} framework. In blue our new similarity head that is able to map similar objects close in the descriptors space. The descriptor can be use for the clustering procedure.}
\label{fig:scheme}

\end{figure*}

The SIMCO two-step algorithm (Fig.~\ref{fig:scheme}) uses a deep architecture which is trained just once, and that generalizes to whichever image.
It is the first multi-class counting by detection approach in this sense: the first step (``Detection'', Sec.~\ref{sec:setup}) provides a single class of generic foreground objects; each detected object is described by a 64-dimensional neural feature vector. The second step (``Clustering'', Sec.~\ref{sec:clustering}) groups the detections into clusters so that each resulting cluster is a different visual ``thing'' with its own count.

\subsection{SIMCO detection} \label{sec:setup}
SIMCO builds upon the Mask-RCNN architecture~\cite{he2017mask} as implemented in~\cite{massa2018mrcnn}, due to its well-known detection capabilities, superior to the previous iterations of the model~\cite{ren2015faster} and with the addition of instance segmentation capabilities. Mask-RCNN is modified to provide, for each detection bounding box, a specific feature descriptor to perform clustering afterwards. Inspired by the work of~\cite{faugeras19833}, we consider each object to be counted as approximable by a particular 2D shape. This assumption is embedded into the detector by training it on a novel 2D shape dataset, \emph{InShape}.

\begin{figure}[tb!]

\begin{minipage}[b]{1.0\linewidth}
  \begin{center}
  %\centerline{\includegraphics[width=8.5cm]{pics/shape_ontology.pdf}}
  \centerline{\includegraphics[width=8.5cm]{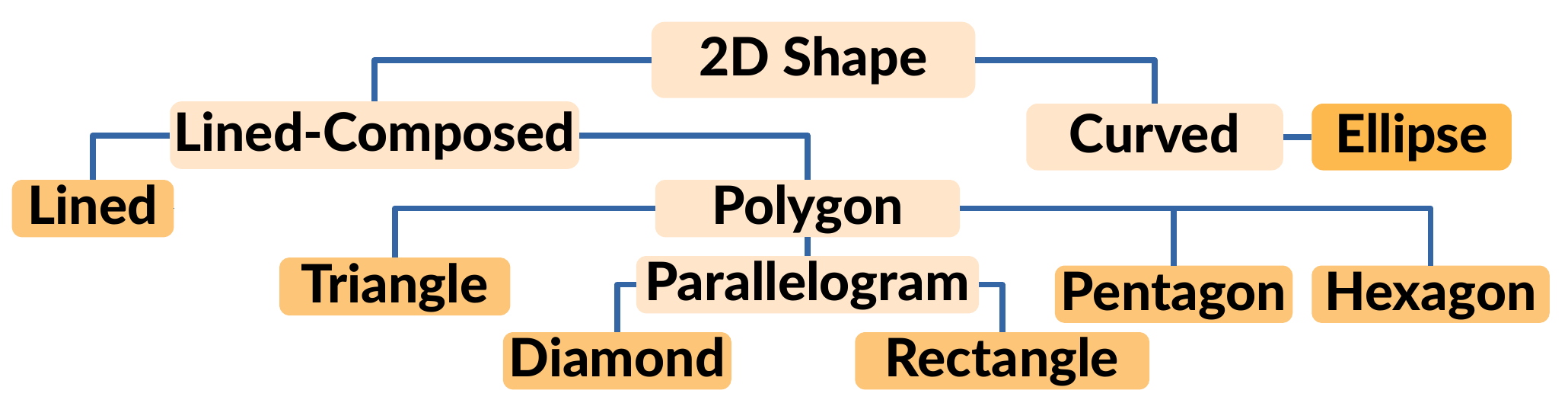}}
   \end{center}
   \vspace{-2em}
  \caption{
2D shape ontology inspired by~\cite{niknam2011modeling}.}
    \label{fig:ontology}
\end{minipage}
\end{figure}

\begin{figure}[tb!]

\begin{minipage}[b]{1.0\linewidth}
  \begin{center}
  \centerline{\includegraphics[width=8.5cm]{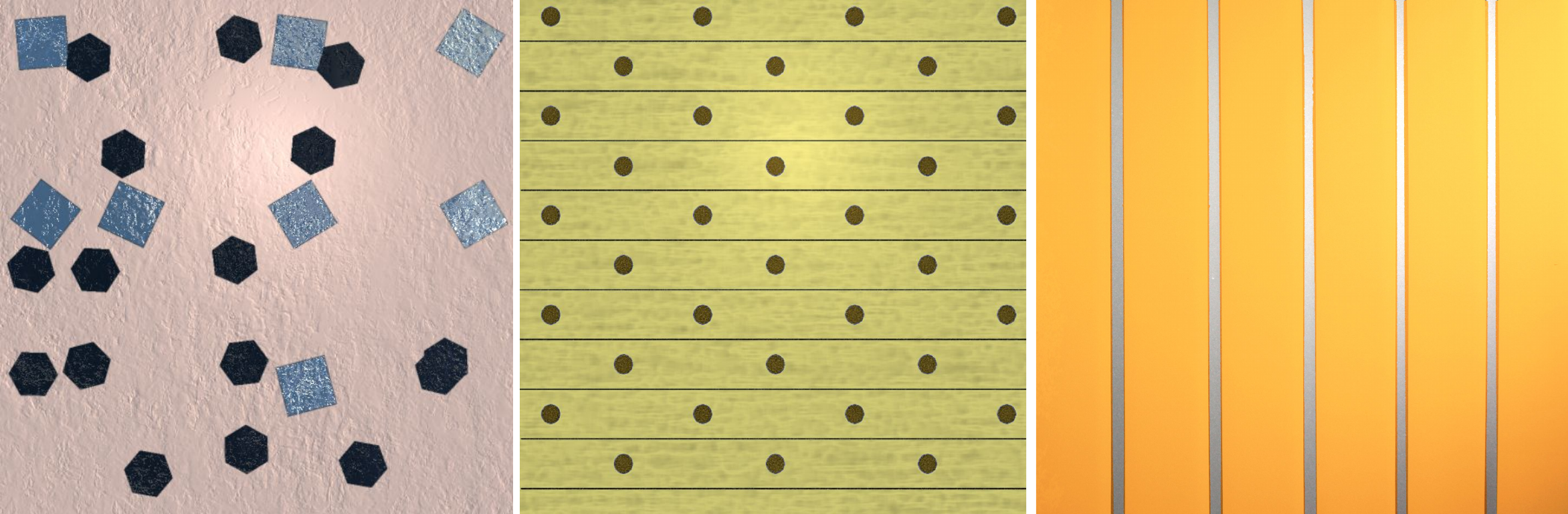}}
  \end{center}
  \vspace{-2em}
  \caption{
\textbf{InShape image samples}. These samples show how different basic shapes are merged together in InShape. In a), two ``types'' of instances are present, dark gray hexagons and light gray squares; in b), circles and lines; in c), lines.}
    \label{fig:dataset}
\end{minipage}
\end{figure}

%\subsubsection{InShape dataset creation}
InShape is a synthetic dataset of 50k images.
Each one contains one or more basic shapes selected from the leaves of a 2D shape ontology inspired by~\cite{niknam2011modeling}: \emph{lined, triangle, rectangle, diamond, pentagon, hexagon, ellipse} (Fig. \ref{fig:ontology}). 
For each shape, we generate instances by varying color (randomly in the RGB space) and scale (randomly from 5\% to 20\% of the image dimension). Instances of different shape classes are co-located in a single image in different spatial patterns (aligned with diverse geometrical layouts, or misaligned and following a Poisson spatial process). Images are made photorealistic (as they were taken from the real world as textural patterns) with the help of the Substance Designer tool (\url{www.allegorithmic.com}). 
Three InShape samples are shown in Fig.~\ref{fig:dataset}. Each image is annotated with bounding box coordinates and labels modeling the shape class, size and color for each object.

%\subsubsection{Similarity branch} \label{sec:simbranch}

Mask-RCNN is modified by adding a new head branch dubbed \emph{similarity head}, providing a similarity-based visual descriptor \(desc(b)\) for each detection bounding box \(b \in D\), where \(D\) is defined as the set of all the detected entities in the current training batch. This branch is placed after the bounding box regression branch of the Mask-RCNN model~\cite{he2017mask}. The \(desc\) function is implemented as a 64-dimensional fully connected layer from the RoI features corresponding to the computed box \(b\). The output of this layer is constrained on the 64-dimensional hypersphere~\cite{ding2015deep} (achieved by normalizing each embedding i.e. \(\|desc(b)\|_{2} = 1\)).

We train this similarity-based descriptor so that: 1) instances of the same class (same basic shape) with the same color and scale (same ''type'') are maximally close; 2) object instances of the same class with different color and/or scale or object instances of different classes have high distance.
These two conditions are formally implemented as a triplet loss \(L_{sim}\), and added to the Mask-RCNN loss as $(L = L_{cls} + L_{box} + L_{mask} + L_{sim})$, with the first three terms being standard~\cite{massa2018mrcnn}, while
the new \(L_{sim}\) is defined as:

\begin{footnotesize}
\noindent
\vspace{1em}
\[\sum_{\substack{(a,p) \in P \\ (a,n) \in N}} max( \|desc(a) - desc(p)\|_{2} - \|desc(a) - desc(n)\|_{2} + \alpha, 0)\]
\end{footnotesize}

Above, \(P\) and \(N\) are respectively the set of positive detection pairs and the set of negative detection pairs, with \(P,N  \subseteq D\times D\); the definition of positive or negative object instance pairs is better described in Sec.~\ref{sec:clustering}. 
%the definition of positive or negative object instance pairs is summarized in Table~\ref{tab:triplet_values}. 
One may note that during training the descriptors of each detection are compared not only within a single image, but also between detections belonging to different images in the same batch, according to the \emph{Batch All} strategy as presented in~\cite{ding2015deep, hermans2017defense}.
The training procedure follows the one presented in~\cite{he2017mask}.
After the training on InShape, SIMCO can be applied to any image with no further fine-tuning. While a domain shift will occur, the purpose of the InShape dataset is to provide a robust and general model that can overcome that as much as possible. For a given test image, it produces a set of detection bounding boxes \(D\) (see Fig.~\ref{fig:qualitative_fish} left), each one equipped with the related feature descriptor \(desc(b)\) (\(b\in D\)).  %In  Fig.~\ref{fig:qualitative_fish} on the left, an example of SIMCO detections.

% \begin{table}[]
% \begin{center}
% \begin{tabular}{|c|c|c||c|}
% \hline

% \textbf{Same Image} & \textbf{Same Class} & \textbf{Same Type} & P/N Pair \\ \hline \hline
% -          & False          & -          & N        \\ \hline
% False          & True          & -          & -        \\ \hline
% True          & True          & False          & N        \\ \hline
% True          & True          & True          & P        \\ \hline
% \end{tabular}
% \end{center}
% \caption{Conditions for \emph{negative} (N) and \emph{positive} (P) pairs are indicated above. ``Same Image'' is true when two objects are in the same image; ``Same Type'' is true when color AND size are the same (``-'' indicates that a value for that relation is not influencing the P/N Pair value). \label{tab:triplet_values} 
% }
% \end{table}

\subsection{SIMCO clustering}  \label{sec:clustering}
The embedding provided by the descriptor  \(desc(b)\) of Sec.~\ref{sec:setup} maps closely objects with the same shape and visual properties so that clustering can be applied to discover natural groupings. The idea is that each cluster is a ``visual thing''. Each cluster represent a class of objects that are distinguished from other classes in the image. The nature of the class depends on the other classes, it could be shape if there are differently shaped objects in the image (e.g. Fig. \ref{fig:dataset}a) or color if all the objects are of the same shape (e.g. Fig. \ref{fig:clusters_qual} bottom row) or a combination of the two (e.g. Fig. \ref{fig:bobode}).
In this phase, outliers that don't belong to any cluster are filtered away (making it possible to count only repeated elements). 

As clustering procedure we choose the affinity propagation algorithm~\cite{frey2007clustering}, since it exploits measures of similarity between pairs of data points (over which we had the Mask-RCNN loss computed) and simultaneously considers all data points as potential exemplars (specific representative of clusters). Affinity propagation has a single parameter (the \emph{preference}) regulating the tendency to select less or more exemplars. 
By varying this parameter one can appreciate how the embedded features organize the detections (Fig.~\ref{fig:qualitative_fish}): starting from a low value and subsequently increasing it, drastically different scales are first separated (Fig.~\ref{fig:qualitative_fish} in the center), followed by shape and color (Fig.~\ref{fig:qualitative_fish} on the right). The correct value of the preference depends on the desired granularity of the clustering. In a real application that depends on the desire of the user. More details on how to chose the parameter for testing are given in the experimental section.

\section{Experiments and Results}
\label{sec:experandresult}

We evaluate counting performance of SIMCO on the Cells~\cite{lempitsky2010learning} and RepTile~\cite{setti2018count} datasets and test it also on novel images to show diverse potential applications. Standard indexes of Mean Absolute Error (MAE) and Normalized Mean Absolute Error (NMAE) are evaluated~\cite{willmott2005advantages}.
These testing datasets are much different from the InShape training dataset, so there is a domain shift between the two, which is expected since the purpose of InShape training was to learn a general and robus description.
%primitive shapes even if they are different from the testing images. 
The challenge in this problem is being able to have good performance of images of any kind without being able to finetune the model on them. A domain shift is always expected and being able to generalize is a very challenging part of agnostic counting.
Datasets such as CLEVR~\cite{johnson2017clevr} or COCO Count ~\cite{chattopadhyay2017counting} are not suitable for our experiments because while they are counting datasets, objects in the image are often not repeated (which is a required assumption for our method) and they belong to a fixed set of classes, so the generalization capabilities of the method would not be tested, just the performance on that particular domain.
In all of the experiments, SIMCO has the Mask-RCNN ResNet50-FPN~\cite{massa2018mrcnn, lin2017feature} as backbone architecture.%, and the preference value of the clustering was set to 1.

In the experiments we compare only with approaches that don't require any kind of training, so agnostic counting approaches such as the one of Lu~\cite{lu2018class} that require a specialization phase are not considered. 

In the experiments with the Cells dataset, we compare with approaches that \emph{don't require any kind of seed initialization}. In the experiments with the RepTile dataset, we compare with approaches that can handle multiple classes by providing seeds for each object class before computation. SIMCO can work in both modalities and so can be compared with different approaches in the two experiments.

\begin{figure*}[t!]
  \begin{center}
  \includegraphics[width=0.98\textwidth]{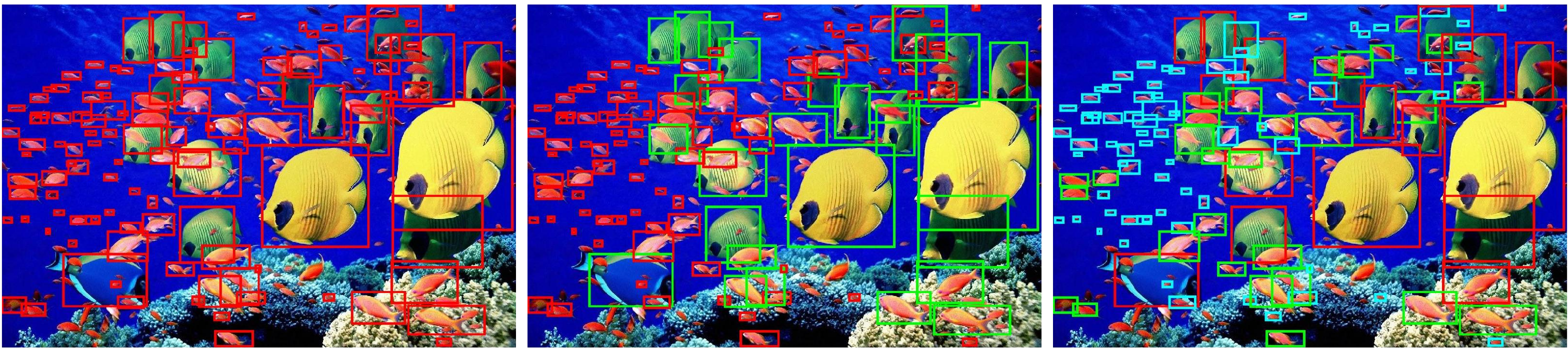}
  \end{center}
  \vspace{-1em}
  \caption{
\textbf{SIMCO clustering process.} By varying the preference value of the affinity propagation clustering, meaningful partitions are obtained. Left: results of the detection (no clustering). Center: the first separation individuates close and far fishes. Right: the second separation captures big circle-like fishes (red), foreground red fishes (green), background red fishes (cyan). It is important to note that while the clustering result changes based on the preference value, none of these results is wrong. A user could want to select each fish regardless of type or select each species separately. The ability to regulate the granularity of the classes is a feature of the method.}
\label{fig:qualitative_fish}

\end{figure*}
\begin{figure}[tb!]

\begin{minipage}[b]{1.0\linewidth}
\begin{center}
\includegraphics[width=0.9\textwidth]{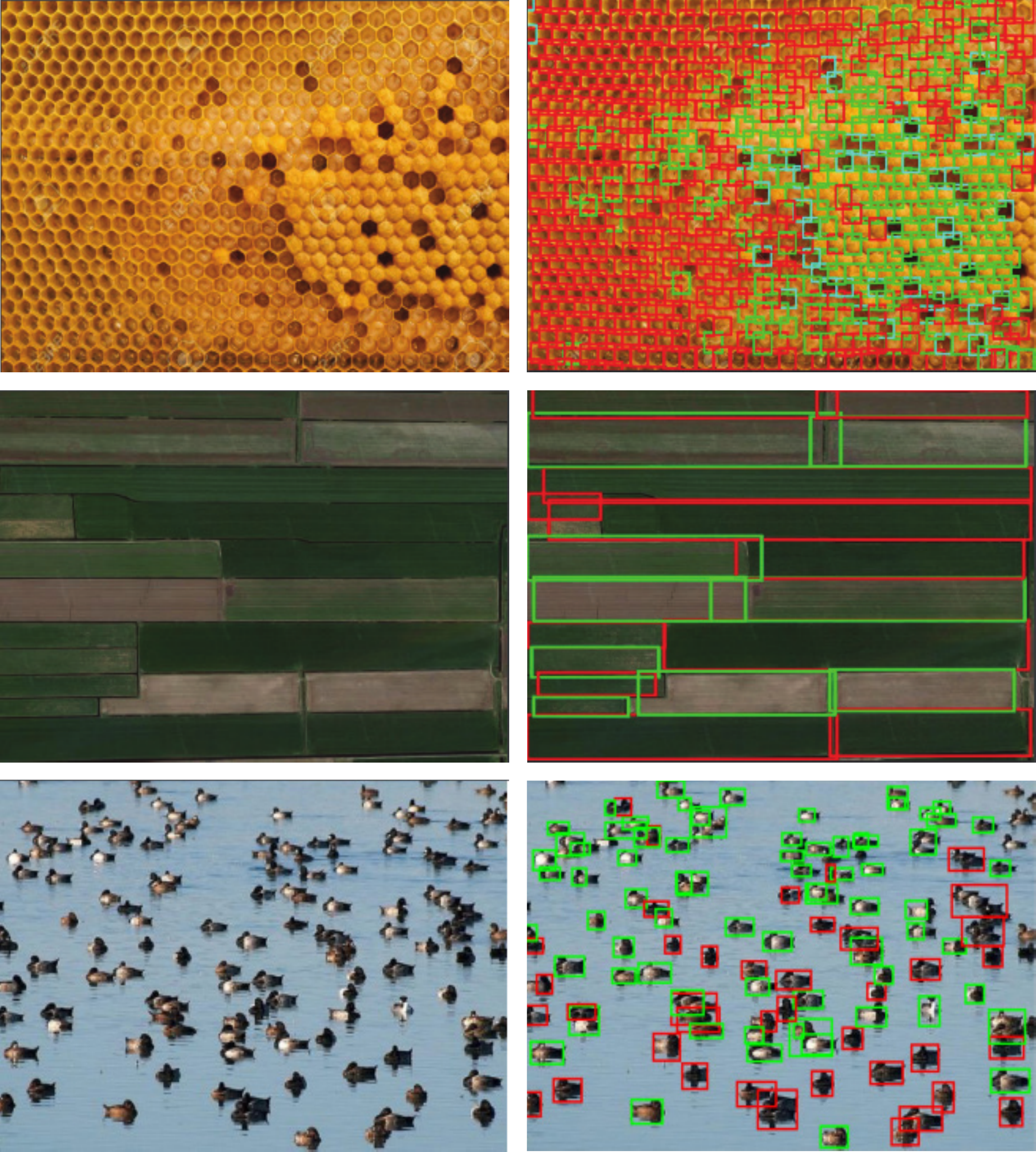}
\end{center}
\vspace{-1em}
\caption{
\textbf{SIMCO on challenging tasks.} Top row: three types of cells are spotted. Middle row: highly elongated shapes such as light/dark green fields are correctly captured. Bottom row: \emph{light and dark belly ducks} are separated into two clusters. This may help an annotator interested only in one of the two species. The high density of the mosaic of bee cells, the difficult shapes ratio in the second task and the fine-grained classes separation in the third further highlight the strengths of SIMCO. }
\label{fig:clusters_qual}
\end{minipage}
\vspace{-0.5em}
\end{figure}
\begin{figure}[tb!]
\begin{minipage}[b]{1.0\linewidth}
  \begin{center}
\includegraphics[width=0.95\textwidth]{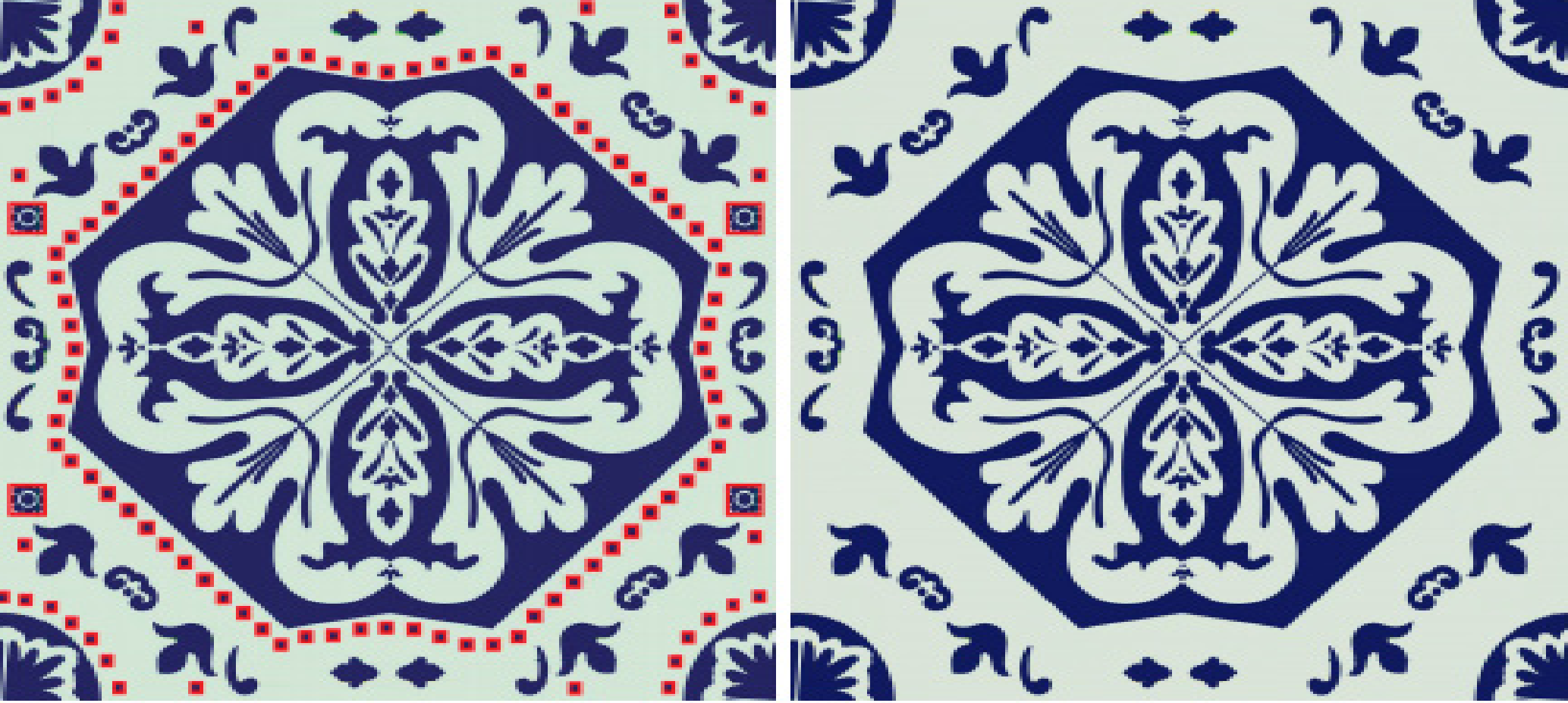}
\end{center}
\vspace{-1em}
\caption{
\textbf{Photoediting task.} A magic wand based on SIMCO can select clusters objects to be removed (dots and flowers on the left). Selected objects can be removed with inpainting (right). 
}
\label{fig:magic_wand_etc}
\end{minipage}
\end{figure}

\textbf{Cells counting.}
The Cells dataset~\cite{lempitsky2010learning} contains images of a single class of cells in challenging spatial configurations (variable density, occlusions) where the detection capabilities of SIMCO can be fully tested.%; therefore, clustering is useless here and the detection capabilities of SIMCO can be fully highlighted.
SIMCO counting performances on the Cells dataset are compared with those obtained by another
%As competitors we consider the 
recent object proposal model that can be used on images \emph{without training on the dataset},  \emph{SharpMask}~\cite{pinheiro2016learning}, and by the \emph{Cai and Baciu}~\cite{cai2013detecting} counting by detection algorithm. These are both single-class general-purpose automated approaches. The SharpMask is generally not a counting approach, but in this experiment it can be used as such since the Cells images contain only one class of elements. By thresholding the proposals provided by the method based on the confidence and then counting how many proposals are found it is possible to use Sharpmask as an effective counting method for a single class of elements \emph{without having ever seen this kind of images} (like our proposed method). The threshold is chosen as the one that gives the best performance in terms of NMAE.
The Cai and Baciu approach follows the setting presented in Setti~\cite{setti2018count}.
To demonstrate the importance of using InShape for training we also report the results obtained training the same Mask-RCNN with COCO images. The same approach of thresholding based on confidence is used with the Mask-RCNN trained on COCO and with SIMCO.

Results are shown in Table~\ref{tab:cells_results}, reporting also the running time on a NVIDIA GeForce GTX 1080 GPU. They show SIMCO definitely performing better than the alternatives.
%ai and Baciu gives the second worst results, 
Results of~\cite{cai2013detecting} are poor due to the high spatial random displacements and the heavy occlusions of the Cells' images; SharpMask scores closer to SIMCO, but behaves badly on occlusions. Mask-RCNN trained on COCO provides the worst counting accuracy scores.

In the table, the results from \cite{lempitsky2010learning} on Cells are not shown as their experimental setting is much easier than ours: those results are obtained after training on a part of the dataset, while the results in the table are obtained \emph{having never seen those images before testing}.

The running time further promotes SIMCO, with one tenth of second per image.

\begin{table}
\begin{center}
\begin{tabular}{|l|l|l|l|l|}
\hline
\textbf{Method} & \multicolumn{2}{l|}{\textbf{Counting}} & \multicolumn{1}{l|}{\textbf{Running}}\\% & \multicolumn{1}{l|}{\textbf{Initialization}}\\
 & MAE & NMAE  & \textbf{Time (s)}  \\ \hline \hline
Cai and Baciu~\cite{cai2013detecting} & 149  & 0.809 & 753\\ \hline
SharpMask~\cite{pinheiro2016learning} & 42 & 0.21 & 8.76 \\ \hline \hline
COCO/Mask-RCNN & 175,65  & 0.99 & 0.12 \\ \hline 
\textbf{SIMCO (no clustering)}  & \textbf{12}  & \textbf{0.07} & \textbf{0.11} \\ \hline
\end{tabular}
 \end{center}
\caption{
\label{tab:cells_results}
 Counting results on Cells~\cite{lempitsky2010learning} dataset. %using automated approaches.
 No clustering is performed since only a single class of elements is expected.
 }

\end{table}
\textbf{SIMCO on RepTile.}
To demonstrate the ability of SIMCO to count unknown objects of multiple classes we tested it on the RepTile dataset~\cite{setti2018count}.  
%RepTile allows to deeply test the generalization properties of SIMCO, since it
RepTile is composed of 50 heterogeneous images (from hyperspectral imagery to aerial photos) with a total of 3173 annotated objects of varying and arbitrarily complex shapes (from candies to airplanes), taken at different scales, illumination conditions etc. In addition, each image has a few specific types of objects annotated while other are not, in order to simulate a user which is interested only in some ``things'' in the image. This makes it even more challenging, as these unannotated objects can be considered distractors that the method needs to deal with.
While the dataset is small, it is very challenging because each image is completely different from the others, so the generalization capabilities of the method can be tested.
This fits perfectly with the clustering parts of SIMCO. In particular, RepTile is thought for a human-in-the-loop approach, since each image has few ``seed'' detections, indicating the types of objects the user is interested to. 
For a fair comparison with the other alternative semi-supervised approaches using seed objects as input, we designed an automatic procedure to set the preference parameter based on the input seeds' annotation. The procedure consists of increasing the preference parameter until each of the seed annotations is covered by a different cluster. Objects of these clusters are counted, while the others are discarded.   

Compared methods are those of Cai and Baciu~\cite{cai2013detecting}, Arteta~\cite{arteta2014interactive} and Setti~\cite{setti2018count}, all exploiting the initial seed annotations (Cai and Baciu can work both automatically as in the Cells test, and with supervision as required by the RepTile protocol).
To demonstrate the advantages of clustering on our similarity branch output, we also report the results obtained with the affinity propagation clustering applied to the standard fully connected feature of the Mask-RCNN classification branch.
Table~\ref{tab:reptile_results_manual} details how SIMCO produces a little more than half of the errors than %the already high performance of 
Setti et al.~\cite{setti2018count}, with significantly lower running time, definitely overcoming all of the other algorithms. Qualitatively, one may appreciate 
Fig.~\ref{fig:qualitative_fish}, showing the process of modulating the clustering until each of the desired seed detection was covered by a cluster; in particular for that image we stop at the second partition, being the yellow and the red fishes requested by the users. 
The same clustering procedure is not as effective when applied to the features of the classification branch (see Table~\ref{tab:reptile_results_manual}).

\begin{figure}[tb!]
\begin{minipage}[b]{1.0\linewidth}
  \begin{center}
\includegraphics[width=8.0cm]{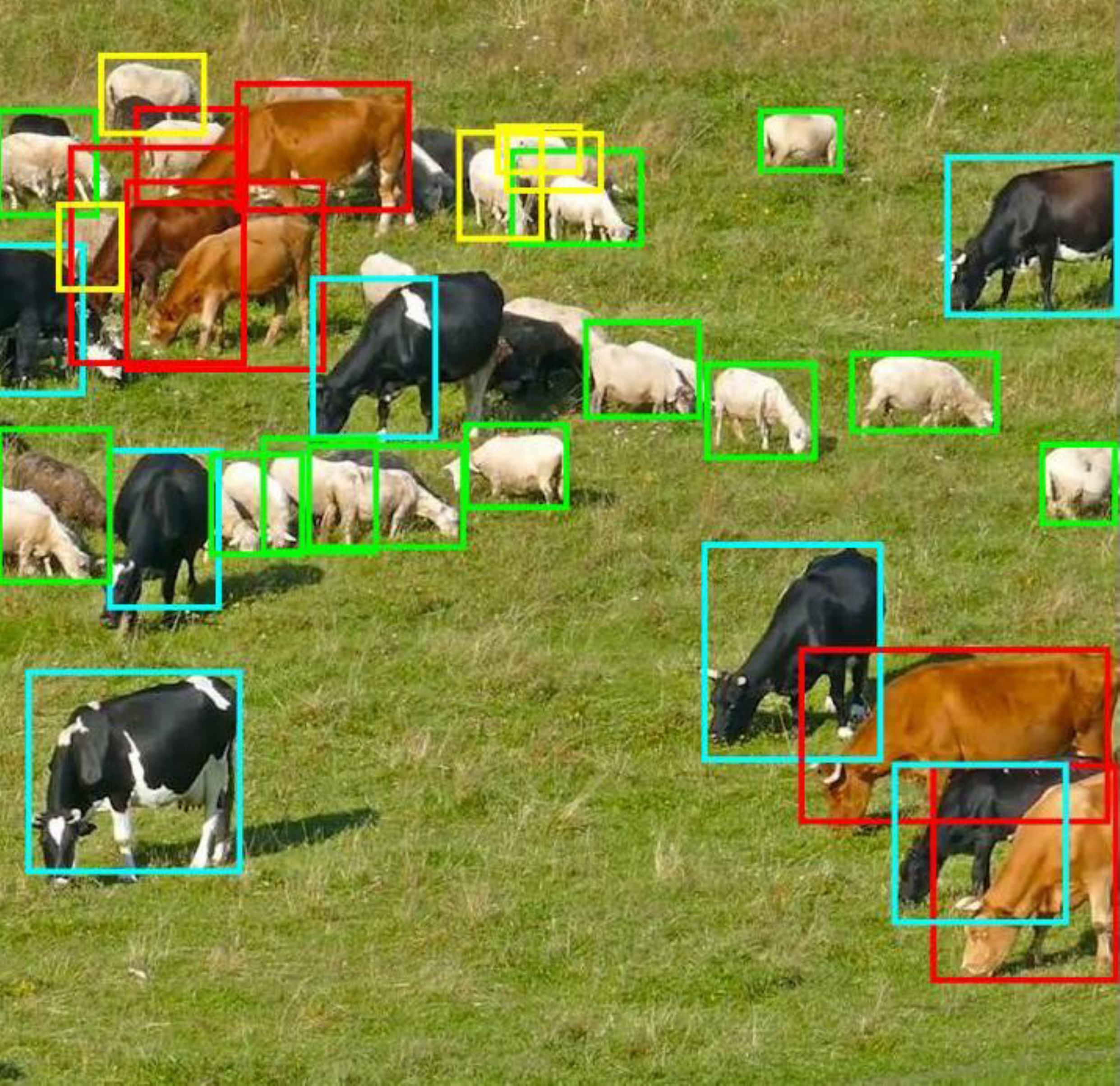}
\end{center}
\vspace{-1em}
\caption{\textbf{Herd animals found by SIMCO.} Clustering is able to not only distinguish between species but also between different varieties of the same type of animal. Cows are in red (for brown cows) or in cyan (for black cows). Green or yellow are sheep.
}
\label{fig:bobode}
\end{minipage}
\end{figure}

\textbf{SIMCO applications.} Due to his high generalization and speed, SIMCO can be the engine for many useful and fancy tasks. SIMCO may help, in a picture, to answer visual questions related to counting, as in the case of Fig.~\ref{fig:lego}. 
In many application domains, it can be used to solve hard multi-class clustering and counting problems, as shown in Figs.~\ref{fig:clusters_qual},~\ref{fig:bobode}.
In top row we see a complex mosaic of three types of bee cells, which SIMCO is able to precisely spot. 
In the middle row we show that SIMCO can handle also highly elongated shapes like fields partitions in remote sensing images, since InShape contains lines.
In the bottom row we show the ability of the method to discriminate fine-grained classes of animals (e.g. \emph{light/dark belly} ducks).
This property can be used, in zoology or other domains, to develop user tools for image annotation aimed at training classifiers for very specific species (e.g., \emph{light belly} ducks).

Finally, a photoediting example is shown in Fig.~\ref{fig:magic_wand_etc}: a magic wand driven by SIMCO can select clusters of similar objects with very few clicks, so that they can be simultaneously removed and inpainted afterwards thanks to the segmentation provided by the underlying Mask-RCNN.

While it could certainly be argued that SIMCO is inferior in raw counting performance to approaches that can be trained and specialized on the specific dataset, it is also true that for many applications (such as a magic wand in photoediting) it is not possible to provide any training set, since only one sample to elaborate may be available. SIMCO is flexible, being able to count with or without any seed initialization. 

\begin{table}[t!]
\begin{center}
\begin{tabular}{|l|l|l|l|}
\hline

\textbf{Method} & \multicolumn{2}{l|}{\textbf{Counting}}& \multicolumn{1}{l|}{\textbf{Running}}\\
 & MAE & NMAE  & \textbf{Time (s)}  \\ \hline \hline
Cai and Baciu~\cite{cai2013detecting} & 59 & 1.034 & 2814\\ \hline
Arteta et al.~\cite{arteta2014interactive}& 50 & 1.629 & 685\\ \hline
Setti et al. TM & 18  & 0.186 & - \\ \hline
Setti et al. TM + CE & 18  & 0.164 & -\\ \hline
Setti et al. complete~\cite{setti2018count} & 14  & 0.109 & 867\\
\hline \hline
COCO/Mask-RCNN/FC & 46  & 0.521 & 0.18\\ \hline
InShape/Mask-RCNN/FC & 19  & 0.272 & 0.18\\ \hline

\textbf{SIMCO}  & \textbf{8.66}  & \textbf{0.086} & \textbf{0.18}\\
\hline
\end{tabular}
 \end{center}
\caption{
\label{tab:reptile_results_manual} Counting results on RepTile~\cite{setti2018count} dataset, exploiting the protocol of~\cite{setti2018count}.}
\end{table}
\vspace{-0.3em}

\section{Conclusion}
\label{sec:conclusion}
We presented SIMCO, a powerful and flexible framework to select and count clusters of similar objects in images. An extensive experimental testing showed that the main ideas behind the method, e.g. training the detection on a custom dataset made of photorealistic images with repeated basic shapes (InShape) and learning an optimal embedding for elements' clustering based on InShape annotations are particularly effective, making the framework suitable for a variety of practical applications in different domains.


\section*{Acknowledgments}
This work is partially supported by the Italian MIUR through PRIN 2017 - Project Grant 20172BH297: I-MALL - improving the customer experience in stores by intelligent computer vision.




\bibliographystyle{IEEEtran}
\bibliography{main}


% \end{thebibliography}

\end{document}